\documentclass[numbers,preprint,review,12pt]{elsarticle}

\usepackage{amssymb}
\usepackage{amsmath}
\usepackage{natbib}

\journal{Pattern Recognition}

\begin{document}

\begin{frontmatter}



\title{A Flexible Recursive Network for Video Stereo Matching Based on Residual Estimation}

\author[label1]{Youchen Zhao}
\ead{20021210736@stu.xidian.edu.cn}
\author[label1]{Guorong Luo}
\ead{22021211823@stu.xidian.edu.cn}
\author[label1]{Hua Zhong\textsuperscript{*}}
\ead{hzhong@mail.xidian.edu.cn}
\author[label1]{Haixiong Li}
\ead{lihaixiong@stu.xidian.edu.cn}

\affiliation[label1]{organization={School of Electronic Engineering, Xidian University},
            addressline={No. 2 South Taibai Road}, 
            city={Xi'an},
            postcode={710071}, 
            state={Shaanxi},
            country={China}}

\begin{abstract}
Due to the high similarity of disparity between consecutive frames in video sequences, the area where disparity changes is defined as the residual map, which can be calculated. Based on this, we propose RecSM, a network based on residual estimation with a flexible recursive structure for video stereo matching. The RecSM network accelerates stereo matching using a Multi-scale Residual Estimation Module (MREM), which employs the temporal context as a reference and rapidly calculates the disparity for the current frame by computing only the residual values between the current and previous frames. To further reduce the error of estimated disparities, we use the Disparity Optimization Module (DOM) and Temporal Attention Module (TAM) to enforce constraints between each module, and together with MREM, form a flexible Stackable Computation Structure (SCS), which allows for the design of different numbers of SCS based on practical scenarios. Experimental results demonstrate that with a stack count of 3, RecSM achieves a 4x speed improvement compared to ACVNet, running at 0.054 seconds based on one NVIDIA RTX 2080TI GPU, with an accuracy decrease of only 0.7\%. Code is available at https://github.com/Y0uchenZ/RecSM. 
\end{abstract}


\begin{highlights}
\item Research highlight 1.

Proposed a stereo matching recursive model for video sequences.
\item Research highlight 2

Promoting convergence based on multi-scale residual estimation.
\item Research highlight 3

A flexible stackable computation structure for different scenarios.
\end{highlights}


\end{frontmatter}


\section{Introduction}
With the remarkable advancement of neural networks, deep learning models have demonstrated unprecedented capabilities in various domains of human life, laying the foundation for the rapid development of computer vision. In fields such as autonomous driving \citep{chen2015deepdriving}, robotics \citep{biswas2011depth}, and others, the maturation of tasks like object detection \citep{fujitake2022temporal, yang2022multi}, tracking \citep{zhang2021siamese, cheng2023tat}, and segmentation \citep{jiang2022super, zhang2023pyramid} has paved the way for practical applications. Stereo matching, as a crucial research area in 3D reconstruction, also holds significant value in these domains.

Stereo matching is a technique aimed at establishing correspondences between a pair of rectified stereo images, resulting in disparities, denoted as 'd', which represent the pixel-wise horizontal offsets in the image coordinate system. These disparities can be further utilized, based on the principles of stereo vision, to estimate depth information in the camera coordinate system \citep{hamzah2016literature}. With the advancement of convolutional neural networks, many deep learning-based stereo matching algorithms have achieved end-to-end matching, significantly reducing matching errors \citep{wang2022uncertainty}. However, existing algorithms are often tailored to specific use cases, making them less adaptable to various scenarios. Some algorithms prioritize high accuracy over real-time performance, which may not be suitable for applications such as autonomous driving and robotics.

In order to perform stereo matching with smaller model parameters and faster processing speed, we introduce a recursive model that calculates only the residuals \citep{wang2019anytime} between the disparities of the previous and current frames, thereby reducing computational complexity. Subsequently, based on the reference frame's disparity map, we recursively compute the disparity results for the current frame to improve the calculation accuracy as much as possible. During the recursion process, we construct a cost volume with temporal attention \citep{zhao2023fast} and a smaller residual search range(\textit{R}), effectively reducing the model parameters and computational load. To enhance the flexibility of the network architecture, we also propose a flexible stackable computation structure to reduce errors, which combined with disparity optimization module, allows for the adjustment of the number of stackable computation structure (SCS) based on specific application requirements, thereby achieving the desired balance between speed and accuracy.

Our main contributions are as follows:

\textbullet We propose a stereo matching recursive model for video sequences, which, with the help of temporal context and temporal attention, successfully presents the current frame disparity result.

\textbullet We propose a multi-scale residual estimation structure based on temporal context, which effectively reduces matching complexity by adapting to changes in the matching range.

\textbullet We propose a flexible stackable computation structure. Leveraging the characteristics of this structure, we iteratively refine the disparity map. With the aid of shared-weight disparity optimization module, we ensure the accuracy of the disparity computation while enhancing the flexibility of the network.

This manuscript is formed into sections as follows. Section \textbf{2} introduces the work related to this paper. Section \textbf{3} introduces the proposed RecSM in detail, including the specific structure of each module. Section \textbf{4} shows the details and performance of the experiment, including ablation experiments of each module and comparison with other existing methods. Finally, Section \textbf{5} describes the conclusions of this paper.
\label{sec:introduction}



\section{Related Work}
Deep learning-based stereo matching algorithms are significantly influenced by the SGM algorithm proposed by Hirschmuller et al. \citep{hirschmuller2007stereo}, and the process is typically divided into four steps: feature extraction, matching cost computation and aggregation, disparity calculation, and disparity refinement. Among these steps, 'matching cost computation and aggregation' directly impact the accuracy of disparity calculation results, while 'disparity refinement' further optimizes the final results. The structural design of these two steps plays a crucial role in the overall network's performance.

Several effective methods for constructing cost volumes have emerged in the literature. Kendall et al. \citep{kendall2017end} used the 'concatenation' method to build a 4D cost volume with a shape of [B, C, D, H, W], where 'D' represents the specified disparity search range. They employed 3D convolution for efficient aggregation. This construction method has provided valuable insights for future stereo matching algorithms. Song et al. \citep{song2019edgestereo} introduced edge cues and jointly constructed cost volumes. Additionally, they reduced computational complexity by computing residual maps at intermediate and large scales. Wu et al. \citep{wu2019semantic} incorporated semantic segmentation as part of the cost volume construction process and built pyramid cost volumes to enrich multi-level semantic and spatial information. Xu et al. \citep{xu2022attention} extracted attention from correlation cues and applied it to the cost volume to suppress redundant information while enhancing matching-related details. Zhao et al. \citep{zhao2023fast} leveraged the previous frame's disparity map as input, generating temporal attention to guide the construction of the correlation-based cost volume, further enhancing the algorithm's speed.

Similarly, many existing algorithms have conducted in-depth research on network structures and post-processing. Wang et al. \citep{wang2019anytime} proposed a staged algorithmic structure that offers flexibility in depth estimation at any desired setting. Khamis et al. \citep{khamis2018stereonet} introduced a disparity refinement module that enables neural network outputs to reach sub-pixel accuracy. Xu et al. \citep{xu2020aanet} proposed the use of 2D deformable convolutions to construct the network, further accelerating inference while maintaining a certain level of accuracy. Zhao et al. \citep{zhao2023fast} employed standard 2D convolutions entirely for network inference, reducing the network's parameter count and making it easy to deploy.

With the development of computer vision technology, stereo matching based on new technologies are gradually showing excellent performance. Tosi et al. \citep{tosi2023nerf} proposed a novel learning framework that utilizes the most advanced NeRF solution to easily train stereo matching networks without the need for any ground-truth. Lou et al. \citep{lou2023elfnet} estimated the evidential-based disparity using transformer and deep evidence learning. Their evidential local-global fusion enables both uncertainty estimation and two-stage information fusion based on evidence, and achieved excellent matching results. Jiang et al. \citep{jiang2024romnistereo} proposed an opposite adaptive weighting scheme that bridges the gap between omnidirectional stereo matching and recurrent all-pairs field transforms, achieving excellent results in the field of panoramic stereo matching research.

In this work, RecSM leverages temporal context as an additional input, computing the disparity results for the current frame solely through residual maps. This approach achieves a smaller search range, reducing network parameter count and accelerating inference speed. Additionally, due to the unique structural characteristics of the stackable computation structure (SCS), it can be stacked to further fine-tune the disparity map. This flexible structure can be applied to a wider range of application scenarios.
\label{sec:Related Work}

\section{RecSM}
\subsection{Network Architecture}
Due to the similarity between consecutive frames in video sequences, we identify areas of change between the images of the preceding and subsequent frames, and only calculate the residuals of these areas and then adjust and compensate them on the disparity map of the previous frame. This approach yields a mathematical model for a recursive structure as follows:
\begin{equation}
\left\{
\begin{split}
\varepsilon_{n, i} &= \mathrm{MREM}_{n, i}(I_{l}, I_{r}, d_{\text{\textit{n-1,i}}}) \\
d_{n, K} &= d_{\text{\textit{n-1,K}}} + \sum_{i=1}^{K} \varepsilon_{n, i}
\end{split}
\right.
\end{equation}
where $\varepsilon_{n, i}$ is the residual result of  Multi-scale Residual Estimation Module (MREM) in \textit{i}-th Stackable Computation Structure($\mathrm{SCS}_i$, \textit{i} = 1...\textit{K}) of frame \textit{n}, $d_{n, K}$ is the disparity result of the $\mathrm{SCS}_K$ in frame \textit{n}, $I_{l}$ and $I_{r}$ are left and right image, respectively. It is worth mentioning that RecSM requires a known disparity map as the initialization value of $d_{0}$. As the algorithm does not require high real-time performance at startup, this disparity map can come from other algorithms that are difficult to meet real-time requirements but have high accuracy.

\begin{figure}[htp]
    \centering
    \includegraphics[width=13cm]{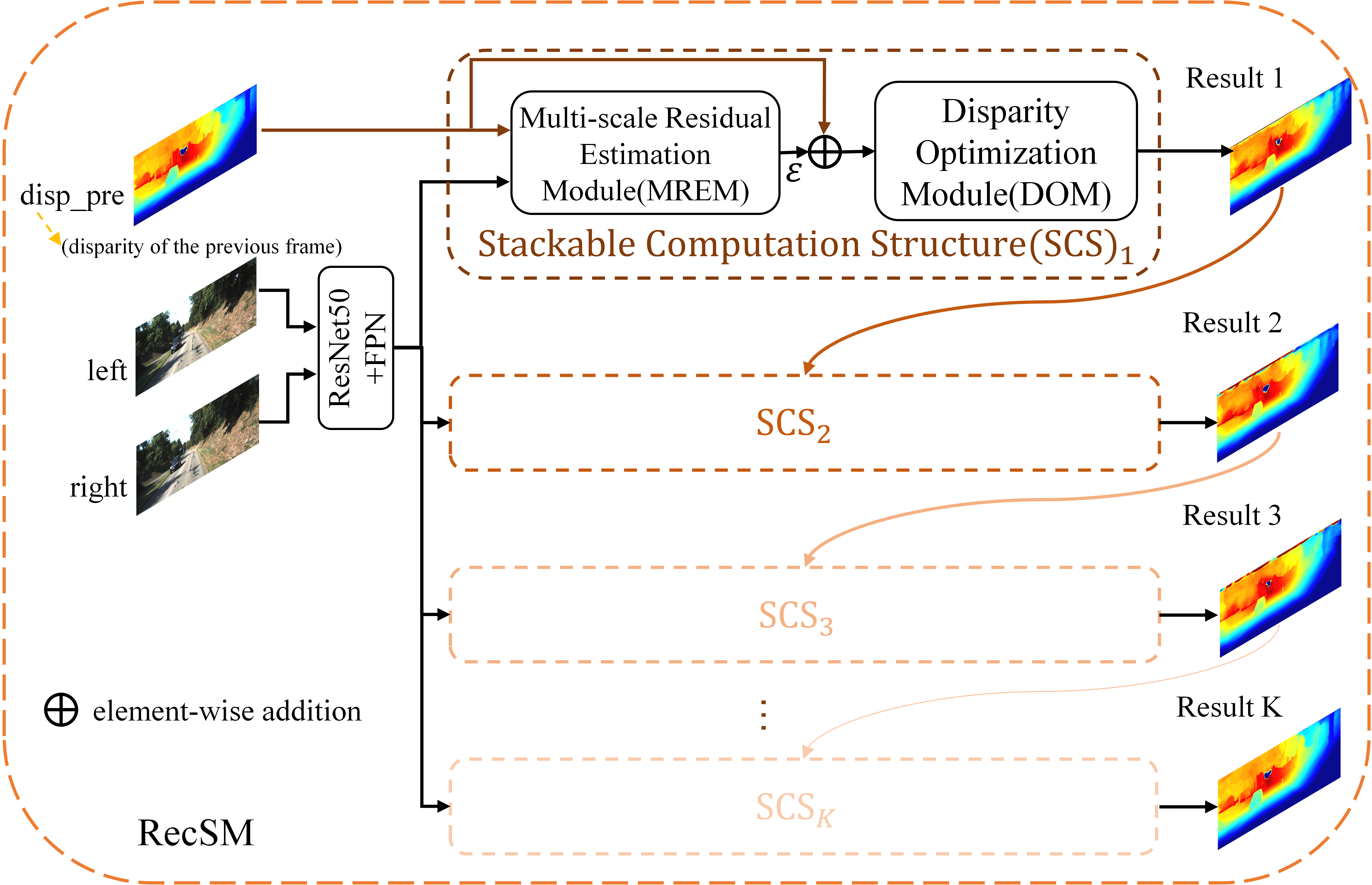}
    \caption{RecSM network architecture.}
    \label{fig:1}
\end{figure}

The complete network structure is shown in Figure \ref{fig:1}, where the inputs consist of the current frame's left and right images, along with the previous frame's disparity map. The output is the disparity map for the current frame's left image.

From top to bottom, there are \textit{K} SCSs in RecSM. For ease of understanding, we will explain RecSM based on \textit{K}=3. The work of Wu et al. \citep{wu2019semantic} and Xu et al. \citep{xu2022attention} validated the effectiveness of ResNet50 \citep{he2016deep} as the feature extraction layer and FPN \citep{lin2017feature} for multi-scale feature extraction in stereo matching, so we continue to use this backbone. Each SCS consists of two components: the MREM and the disparity optimization module(DOM), which together form a SCS. Inside the MREM, residuals $\varepsilon$ are computed on each scale branch, and temporal attention is fused for searching, resulting in the disparity map for the current frame. The difference in input between each module lies in the temporal context. $\mathrm{SCS}_\textit{1}$ takes the previous frame's disparity map as its temporal context, while subsequent SCSs take the disparity results from the previous module as their temporal context. The outputs of MREM are refined by DOM with shared weights, yielding the final output for each SCS.

\subsection{Temporal Context-Based Residual Estimation Module}
\subsubsection{Temporal Context}
\begin{figure}[htp]
    \centering
    \includegraphics[width=13cm]{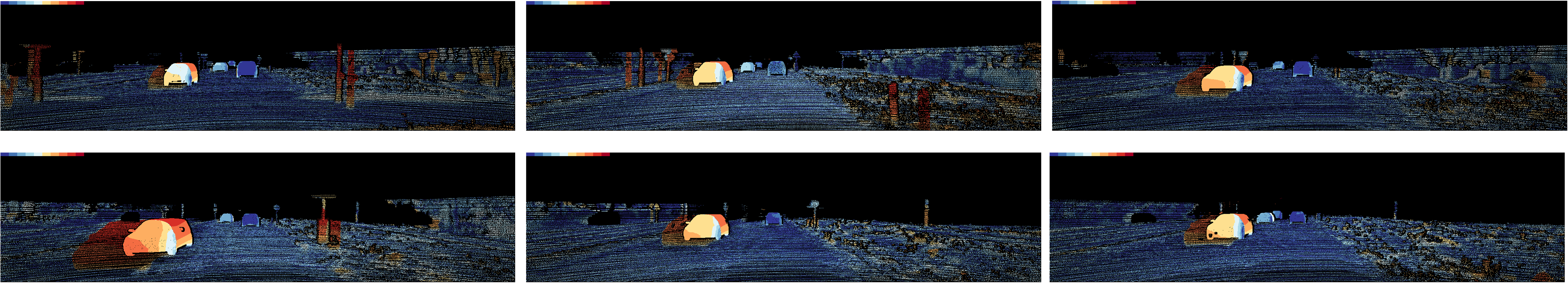}
    \caption{Visualization of disparity changes in continuous frames of road scenes. Warmer colors indicate larger changes, while cooler colors represent smaller variations.}
    \label{fig:2}
\end{figure}

In scenarios such as autonomous driving and robotics, there is a high degree of similarity between disparity maps of consecutive frames \citep{zhao2023fast}. This allows for the allocation of computational resources to regions with significant disparities between the disparity maps of the previous and subsequent frames. Figure \ref{fig:2} provides a visualization of the variation differences in disparity maps between consecutive frames for certain scenes in the KITTI 2015 dataset \citep{menze2015object}. It can be observed that regions with significant disparity variations across the entire road scene correspond to objects such as vehicles, pedestrians, and landmarks. Moreover, as the distance between the vehicle and the camera decreases, the areas with significant disparity changes expand, and the disparity variation range increases. Similarly, although disparity variation regions exist on objects, the variation at the center of the objects is relatively small.

\begin{figure}[htp]
    \centering
    \includegraphics[width=13cm]{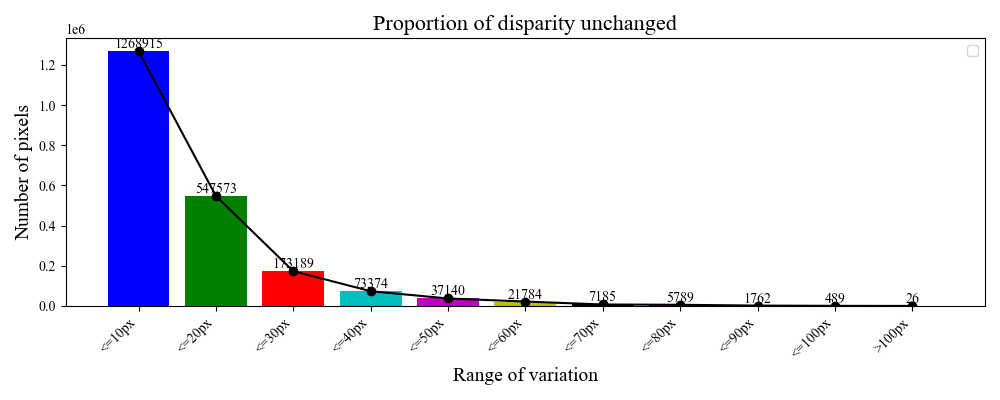}
    \caption{Visualization of disparity change distribution in road scene.}
    \label{fig:3}
\end{figure}

Observing the specific numerical distribution of areas with significant disparity changes in the KITTI 2015 dataset, after conducting statistical analysis, it was found that the disparity change range for road scenes falls between -104 and 208 (detailed dataset usage instructions will be provided in Section 4.1). The distribution of disparity changes exceeding 3 pixels is illustrated in Figure \ref{fig:3}.

59.37\% of the disparity changes fall within the range of (3px, 10px], 25.62\% within (10px, 20px], and 8.1\% within (20px, 30px]. This suggests that in road scene scenarios, most disparity change values remain within a relatively narrow interval, which corresponds to the \textit{R}. Compared to a disparity search range of 192, using temporal context to calculate only residuals can significantly reduce the search range and computational workload.

\subsubsection{Multi-Scale Residual Estimation Module}
\begin{figure}[htp]
    \centering
    \includegraphics[width=13cm]{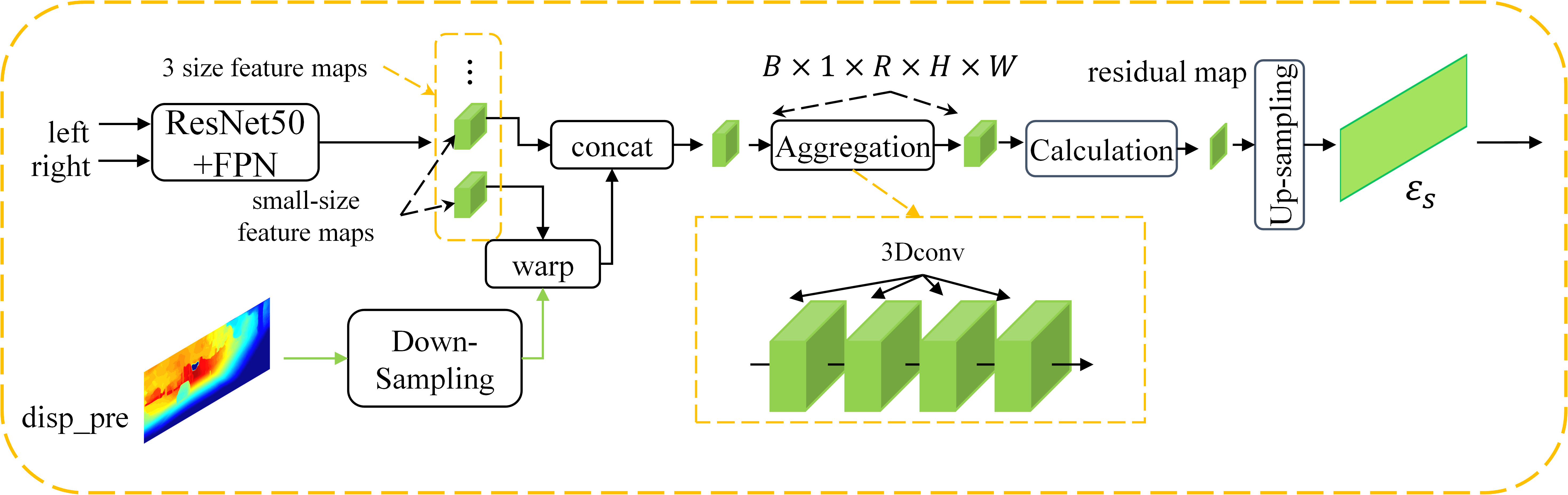}
    \caption{Single-Scale residual estimation module (small-scale).}
    \label{fig:4}
\end{figure}

Due to the fact that differences in consecutive frame disparities mainly manifest in the details of moving objects, calculating residuals directly in areas with image details such as edges and contours can be challenging. However, the changes in the central regions of the background and objects are relatively small. Therefore, initial residual estimation on smaller-scale feature maps with larger receptive fields can alleviate the computational difficulty. Subsequently, through hierarchical fusion of residual results calculated layer by layer for image detail regions, the small-scale residual estimation module is depicted in Figure \ref{fig:4}.

\begin{figure}[htp]
    \centering
    \includegraphics[width=13cm]{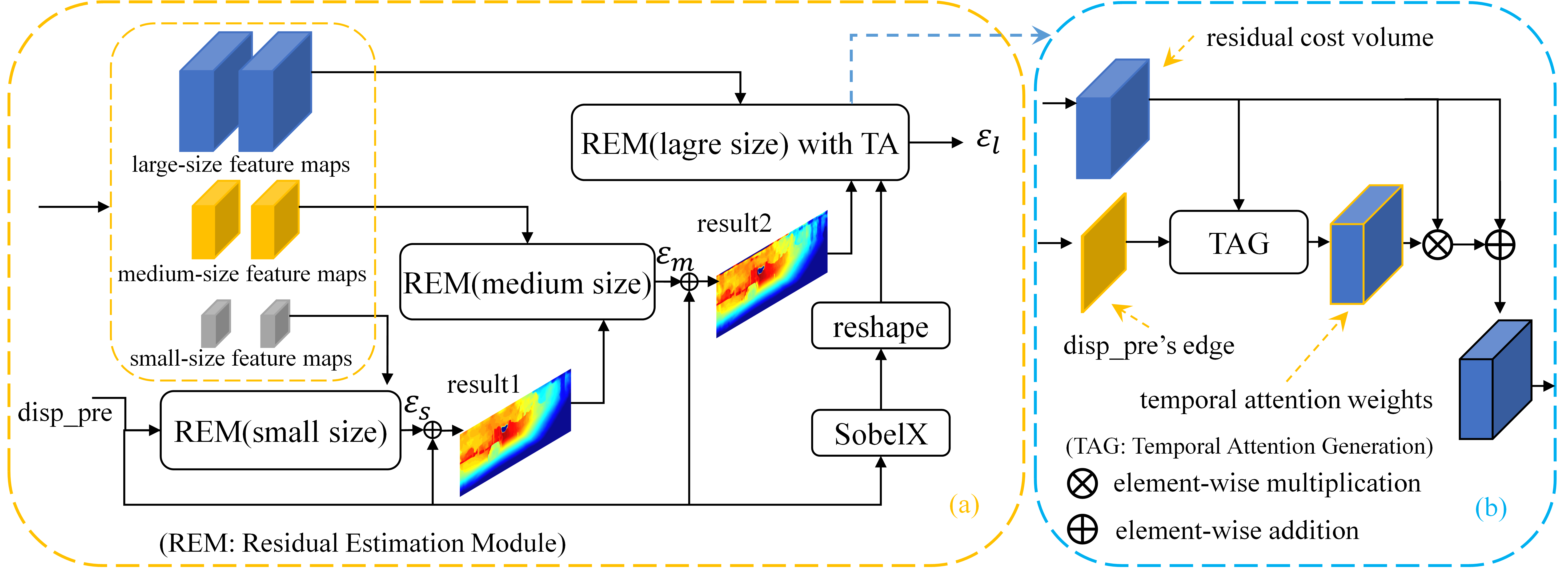}
    \caption{(a) MREM (b) Temporal attention fusion structure in the large-scale branch.}
    \label{fig:5}
\end{figure}

After obtaining the disparity computation result from the small-scale branch, it used as input for the computation in the medium-scale branch. Subsequently, the output from the medium-scale branch serves as input for the large-scale branch in a progressive manner to achieve a more accurate disparity result. During the computation in the large-scale branch, we employ temporal attention generated from the edges of the previous frame's disparity map. This temporal attention is applied to the residual cost volume of the large-scale branch to enrich fine-detail feature information. The entire MREM is illustrated in Figure \ref{fig:5}(a), and the functioning of temporal attention in the large-scale branch is depicted in Figure \ref{fig:5}(b).

\subsection{Stackable Computation Structure}
\subsubsection{Structure of SCS}
Inspired by the work of Wang et al. \citep{wang2019anytime}, we utilize MREM and disparity optimization module (DOM) to construct a flexible and highly versatile structure that can be applied to algorithms in various application scenarios. This approach aims to strike a balance between speed and accuracy and is referred to as the SCS. The specific structure is illustrated in Figure \ref{fig:1}.

For one SCS, the input comprises the current frame's left and right images along with a disparity map used for residual estimation. The output is the disparity map for the current scene. Notably, both the input and output consist of disparity maps. Leveraging this structural characteristic, disparity maps can be recursively fed into multiple SCSs. 

We have stacked a maximum of 3 SCSs in our approach. In theory, with sufficient hardware resources and computational power, the number of SCSs could be increased. As the number of SCSs increases, the error decreases, but the computational time also increases. In practical usage, we can adjust the number of SCSs based on the requirements of the specific scenario. If real-time performance is a priority and high accuracy is not critical, we can use fewer SCSs. On the other hand, if high accuracy in disparity calculation is crucial for the scenario, we can increase the number of SCSs to meet the accuracy requirements.

\subsubsection{Disparity Optimization Module}
\begin{figure}[htp]
    \centering
    \includegraphics[width=13cm]{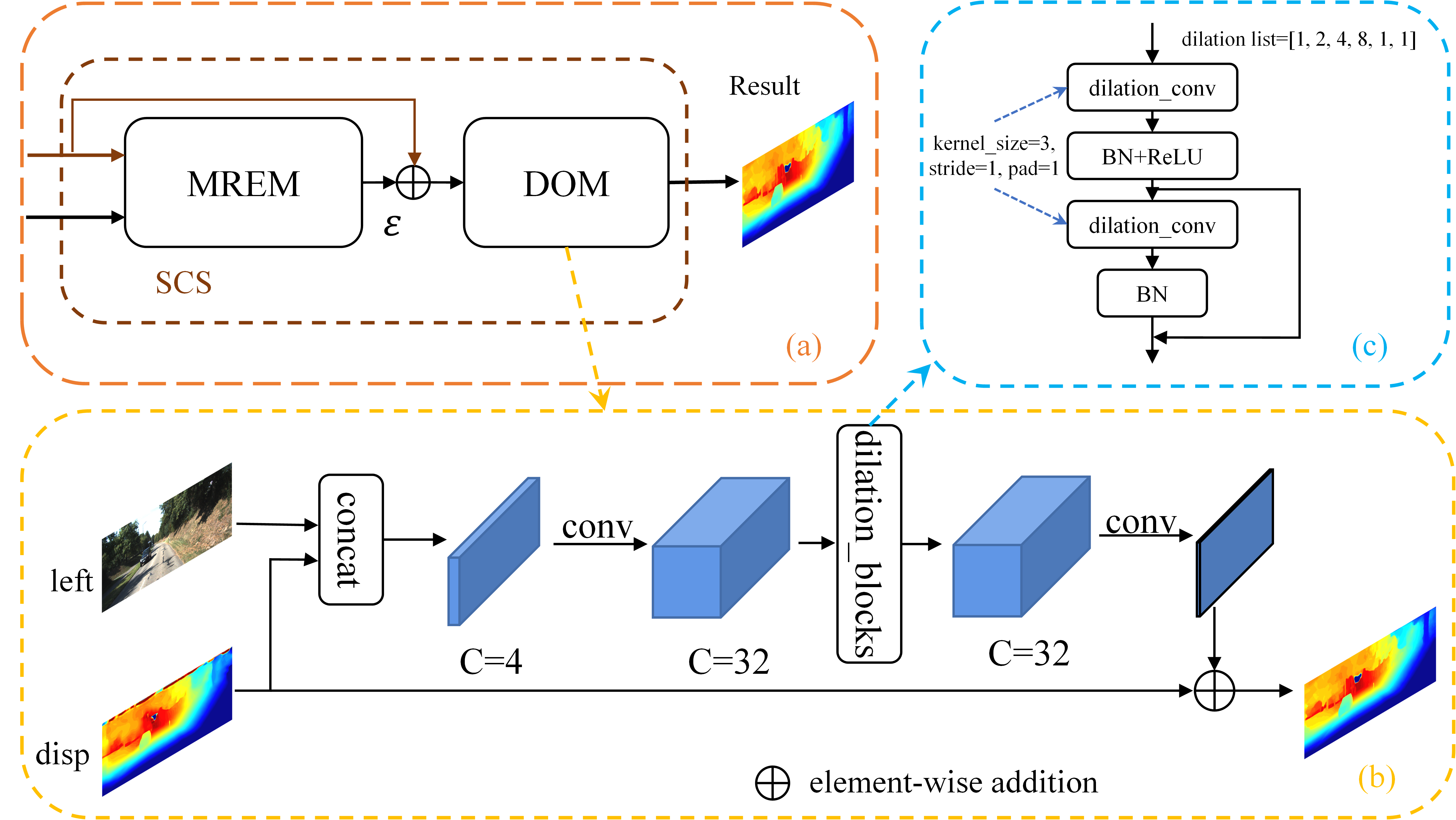}
    \caption{(a) Single SCS. (b) DOM. (c) Dilated Convolution Module Structure.}
    \label{fig:6}
\end{figure}

When performing disparity recovery solely through the calculation of residual results, there are certain limitations due to the absence of direct computation of the matching information between the left and right images. Inspired by the work of Khamis et al. \citep{khamis2018stereonet} and Xu et al. \citep{xu2020aanet}, we incorporate a shared-weight DOM after each MREM to achieve disparity optimization. The specific structure is illustrated in Figure \ref{fig:6}.
\label{sec:RecSM}

\section{Experiment}
\subsection{Datasets and evaluation metrics}
KITTI stereo evaluation includes KITTI 2012 \citep{geiger2012we} and KITTI 2015 \citep{menze2015object}. KITTI 2012 contains 194 training and 195 testing image pairs, and KITTI 2015 contains 200 training and 200 testing image pairs. The resolution of KITTI 2015 is 1242$\times$375, and that of KITTI 2012 is 1226$\times$370. Since KITTI only provides ground-truth disparity maps for the 10th frame in each scene, we used PSMNet \citep{chang2018pyramid} to calculate the disparity maps of the previous frame as additional data for training.

KITTI stereo evaluation commonly used three evaluation metrics, which are End-point Error (EpE), percentage of disparity outliers (D1-all), and Run-time. This paper will evaluate the algorithm performance based on these three metrics.

\subsection{Implementation details}
The RecSM architecture was implemented using PyTorch 1.7. All models were end-to-end trained with Adam \citep{kingma2014adam} ($\beta_{\text{1}} = 0.9$, $\beta_{\text{2}} = 0.999$). During training, images were randomly cropped to size H = 256 and W = 512. The batch size was set to 4 for the training on one NVIDIA RTX 2080TI GPU. The setting of learning rate (lr) refers to the idea of Li et al. \citep{li2022practical}. We use 10 epochs from the beginning of training to warm up. The lr increases linearly from 5.8e-5 to 4e-4, then maintains the training learning rate unchanged to the 300th epoch, and then decreases linearly to 2e-6 at the 700th epoch.

For the $\mathrm{SCS}_K$, the calculation of the total loss function $loss_{\textit{total}}$ is as follows:
\begin{equation}
loss_{\textit{total}} = \sum_{i=1}^{K} \lambda_{i} * loss_{i}
\end{equation}

The calculation of the loss function $loss_{i}$ for each $\mathrm{SCS}_i$ is as follows:
\begin{equation}
loss_{i} = 0.5 * loss_s + 0.7 * loss_m + 0.9 * loss_l
\end{equation}

The loss function for each scale branch is computed using the smooth L1 loss.

\subsection{Ablation Study}
\subsubsection{Analysis of MREM}
\begin{figure}[htp]
    \centering
    \includegraphics[width=13cm]{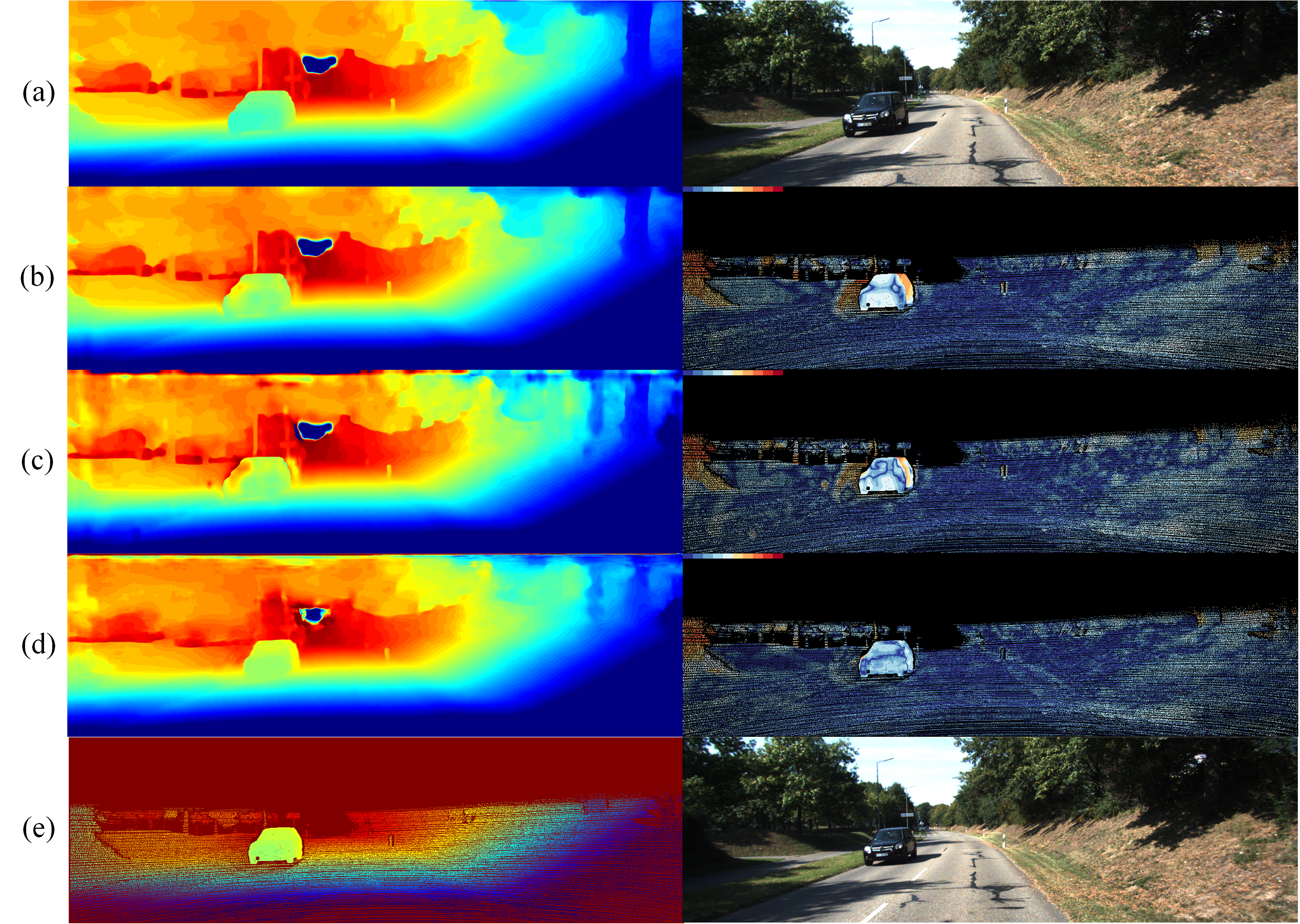}
    \caption{Visualization of outputs from the MREM. (a) Left: Disparity map from the previous frame; Right: Original left image from the previous frame. (b) Left: Disparity map output from the small-scale network branch; Right: Visualization of error map computed against the corresponding ground-truth using occlusion (occ) labeling. (c) Left: Disparity map output from the medium-scale network branch; Right: Visualization of error map. (d) Left: Disparity map output from the large-scale network branch; Right: Visualization of error map. (e) Left: Ground-Truth disparity results; Right: Current frame's left image.}
    \label{fig:7}
\end{figure}

MREM consists of three branches with different scales, and the internal residual output of each scale is added to the disparity to obtain the results of each branch. The error situation is shown in Table \ref{tab:1}. In order to visually compare and observe the output results of each scale, the error visualization results of the three scales are shown in Figure \ref{fig:7}.
\begin{table}[htbp]
\centering
\caption{Comparison of output results at different scales within MREM. }
\begin{tabular}{c|cc}
\hline
Scale in MREM & EpE(px) & D1-all(\%) \\
\hline
 Small-scale & 2.176 & 15.25 \\
 Medium-scale & 1.312 & 8.003 \\
 Large-scale & 0.802 & 3.082 \\
\hline
\end{tabular}
\label{tab:1}
\end{table}

\begin{figure}[htp]
    \centering
    \includegraphics[width=13cm]{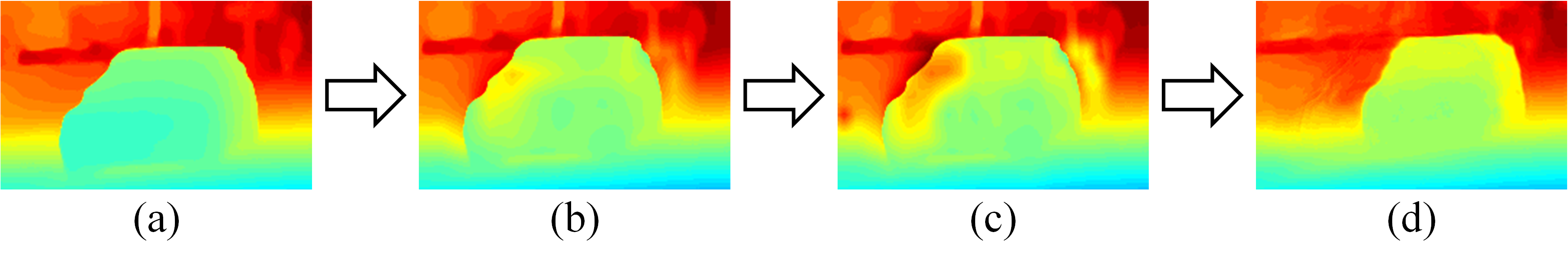}
    \caption{Disparity calculation results for automotive target regions in the image. (a) Disparity map from the previous frame. (b) Output from the small-scale network branch. (c) Output from the medium-scale network branch. (d) Output from the large-scale network branch.}
    \label{fig:8}
\end{figure}

We primarily observe the layered changes in disparity within the automotive target regions, as shown in Figure \ref{fig:8}.

\begin{figure}[htp]
    \centering
    \includegraphics[width=7cm]{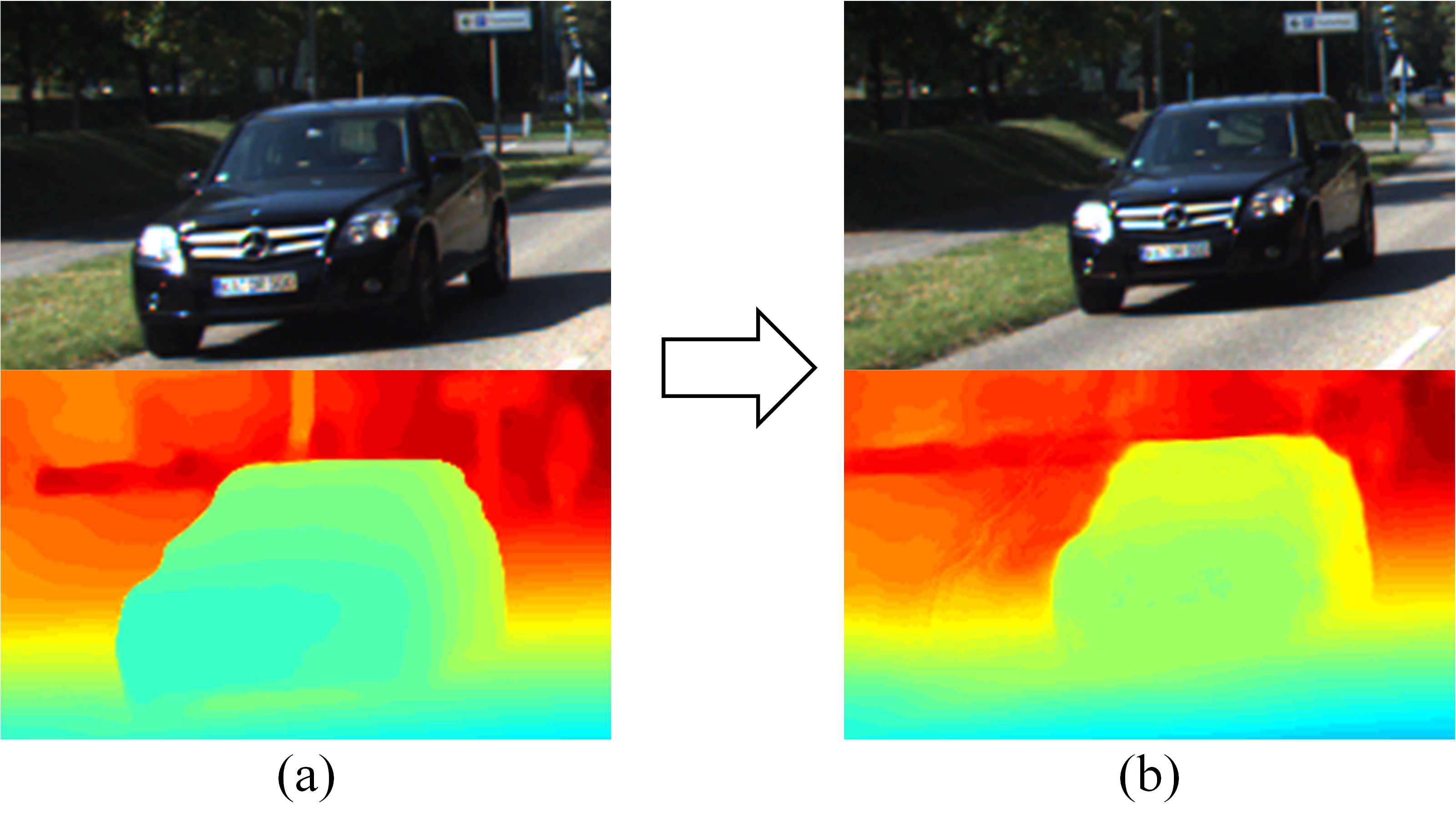}
    \caption{Comparison of the car object in the original image and disparity map within the image. (a) Previous frame. (b) Current frame.}
    \label{fig:9}
\end{figure}

In Figure \ref{fig:8}, (b) compared to (a), there is an overall change in the disparity for the car. However, due to the limited image details contained in the small-scale feature map, the areas such as edges and contours in the disparity result still require adjustments. With the medium-scale branch, (c) shows that adjustments have begun in the edge and contour areas. The disparity in the changing region between the left and right frames on the left side gradually decrease, transitioning into the background, while the disparity on the right side of the region start to increase, gradually recovering the disparity for the right part of the car in the current frame. After passing through the large-scale branch, (d) shows that the car's contours match the contours of the car in the original image of the current frame closely, and the disparity has been successfully propagated. The comparison of the disparity maps between the previous and current frames with the original image is presented in Figure \ref{fig:9}.

In MREM, large-scale branches fuse temporal attention, which further improves the computational efficiency of residuals. When stacking 1 SCS, the comparison of TA ablation experiments is shown in the Table \ref{tab:2}.
\begin{table}[htbp]
\centering
\caption{Comparison of experimental metrics using TA for large-scale branching in MREM. (\textbf{bold} indicates best)}
\begin{tabular}{c|cc}
\hline
Use TA & EpE(px) & D1-all(\%) \\
\hline
  & 0.981 & 4.467 \\
\checkmark & \textbf{0.802} & \textbf{3.089} \\
\hline
\end{tabular}
\label{tab:2}
\end{table}

It can be observed that the disparity map for the current frame undergoes adjustments after passing through the MREM, resulting in the propagation of the disparity for the current frame.

\subsubsection{Analysis of SCS}
Considering that $\mathrm{SCS}_\textit{1}$ has already computed the disparity map for the current frame, subsequent $\mathrm{SCS}_i$ can be viewed as further optimization steps. Therefore, the \textit{R} in the subsequent $\mathrm{SCS}_i$ can be moderately reduced compared to the $\mathrm{SCS}_\textit{1}$. We employ a dynamic setting approach where the \textit{R} gradually decreases with each additional SCS. When the number of SCSs is set to 3, the specific configuration is detailed in Table \ref{tab:3}.
\begin{table}[htbp]
\centering
\caption{Dynamic setting of \textit{R} at different scales with the number of SCSs}
\begin{tabular}{c|ccc}
\hline
$\mathrm{SCS}_i$ & $\mathrm{SCS}_\textit{1}$ & $\mathrm{SCS}_\textit{2}$ & $\mathrm{SCS}_\textit{3}$ \\
\hline
 Large-scale \textit{R} & 16 & 8 & 4 \\
 Medium-scale \textit{R} & 8 & 4 & 2 \\
 Small-scale \textit{R} & 4 & 2 & 1 \\
\hline
\end{tabular}
\label{tab:3}
\end{table}

Compared to using the same \textit{R} for each $\mathrm{SCS}_i$, our approach aligns more with the intended use of SCS. It not only improves computational speed but also reduces mismatches caused by excessively large \textit{R}. The comparison of metrics for dynamically setting the \textit{R} is presented in Table \ref{tab:4}.
\begin{table}[htbp]
\centering
\caption{Comparison of experimental metrics corresponding to the dynamic setting of \textit{R}.}
\begin{tabular}{c|ccc}
\hline
Whether the \textit{R} Varies & EpE(px) & D1-all(\%) & Run-time(s) \\
\hline
   & 0.896 & 3.284 & 0.077 \\
 \checkmark & \textbf{0.737} & \textbf{2.336} & \textbf{0.054} \\
\hline
\end{tabular}
\label{tab:4}
\end{table}

\begin{figure}[htp]
    \centering
    \includegraphics[width=13cm]{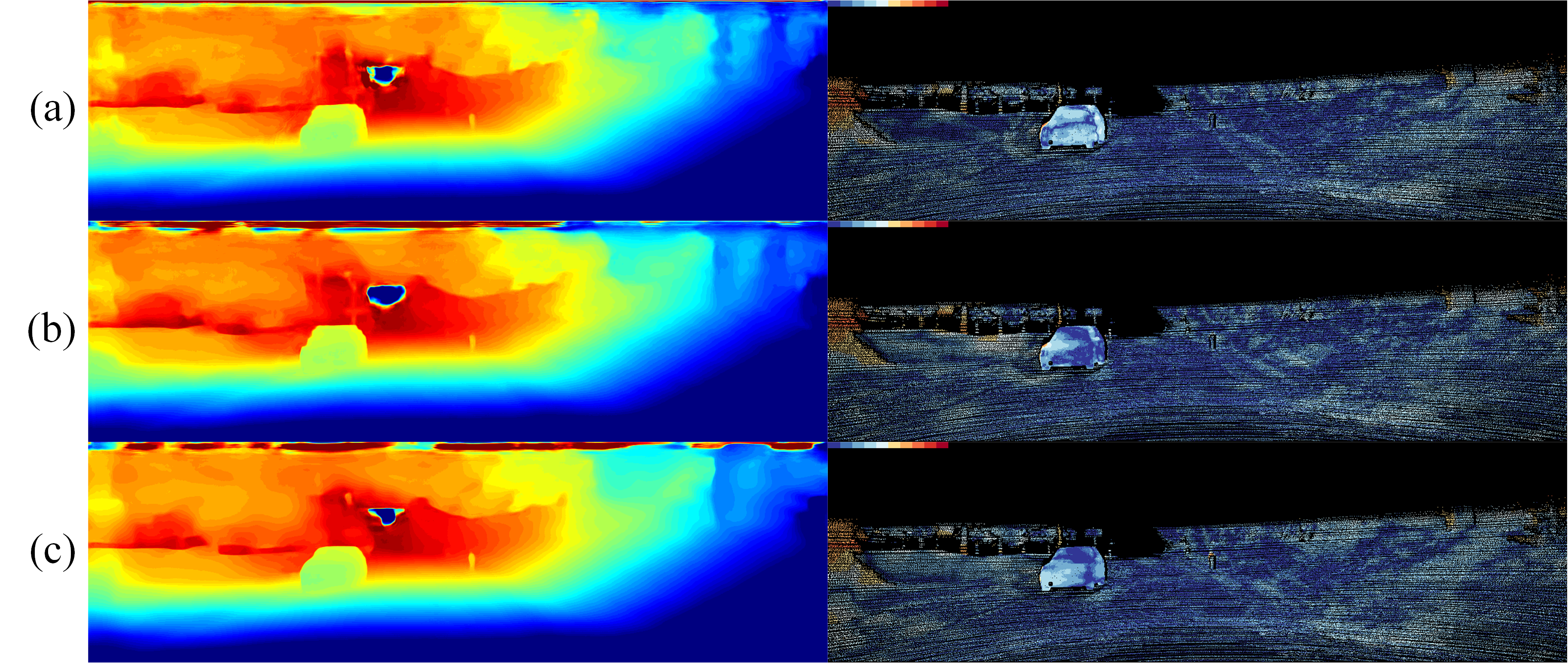}
    \caption{Output disparity maps and error visualization results for different numbers of SCSs. (a) 1 SCS. (b) 2 SCSs. (c) 3 SCSs.}
    \label{fig:10}
\end{figure}

The output disparity maps and error visualization results for each SCS are shown in Figure \ref{fig:10}. As the number of SCS in the stack increases, the errors gradually decrease, and the computational effectiveness in image detail regions improves progressively.

\begin{figure}[hp]
    \centering
    \includegraphics[width=13cm]{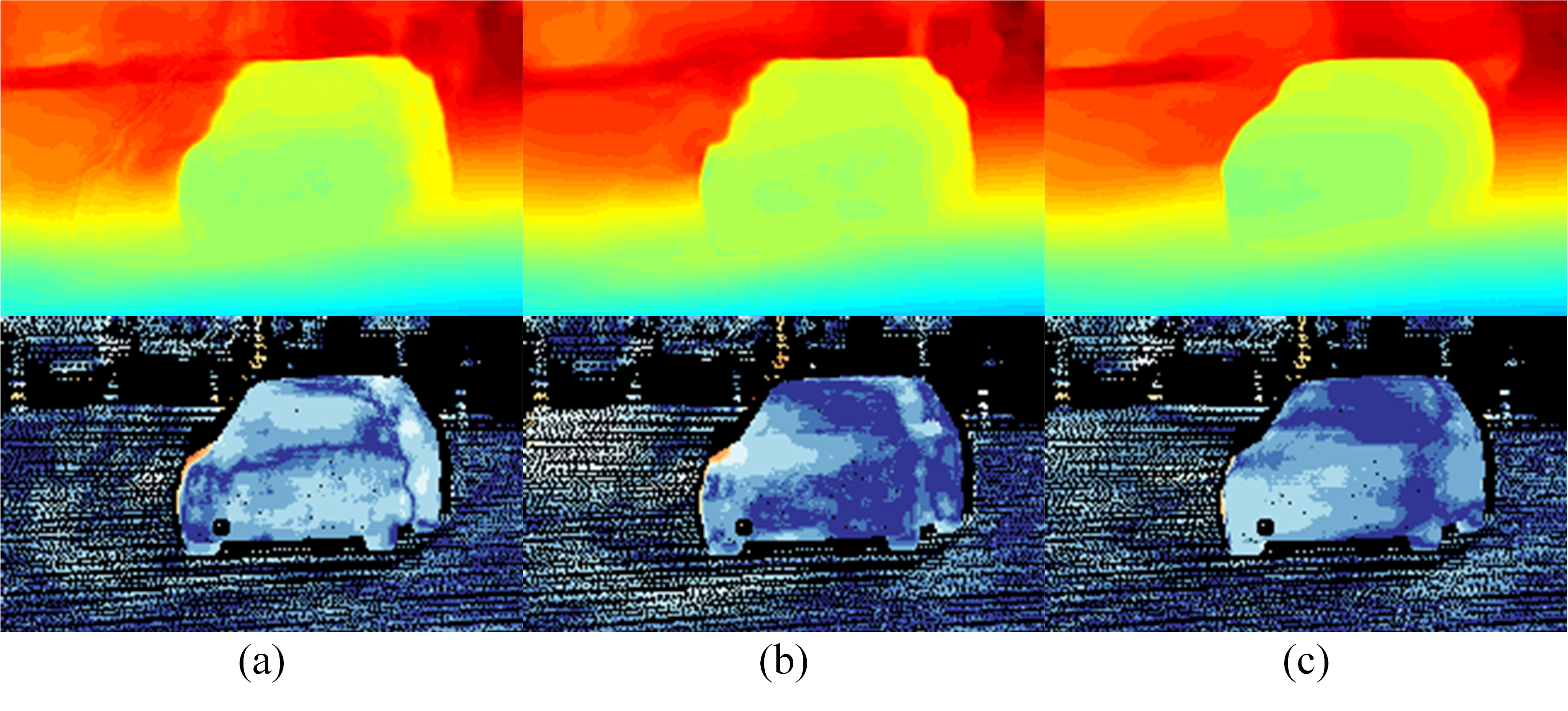}
    \caption{Disparity calculation and error visualization results for the automotive target region in the image with different numbers of SCSs. (a) 1 SCS. (b) 2 SCSs. (c) 3 SCSs.}
    \label{fig:11}
\end{figure}

When examining the calculation results for the automotive target region with varying numbers of SCS, as depicted in Figure \ref{fig:11}, it is observed that with an increasing number of SCSs, the disparity calculation error in the image detail regions gradually decreases. Slight discrepancies are noticed in (a) and (b), where the contours of the car from the previous frame were not adjusted correctly. These errors are rectified in (c), where the car's region not only exhibits smoother disparity map contours but also lower errors compared to (a) and (b).

\begin{figure}[htp]
    \centering
    \includegraphics[width=13cm]{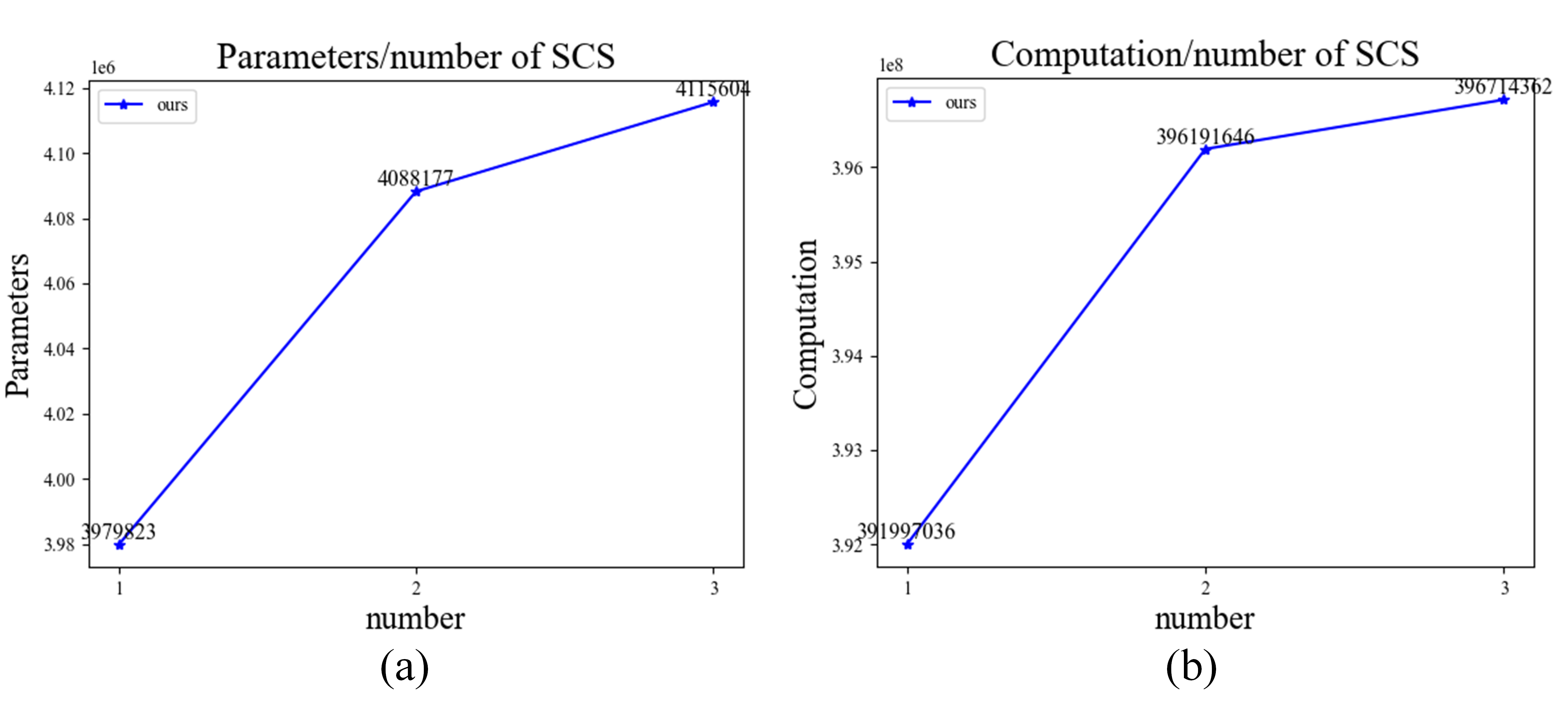}
    \caption{The number of parameters and computation complexity of the network when stacking different numbers of SCS. (a) Parameters. (b) Computation.}
    \label{fig:12}
\end{figure}

When stacking different numbers of SCS, the corresponding computational and parameter quantities of the network are shown in the Figure \ref{fig:12}. As the number of module stacks increases, the computation and parameters of the network also increase accordingly. However, due to the clever structural design of the SCS module and the unique advantages of residual calculation, the increase in parameters and computation is slower. 

\begin{figure}[htp]
    \centering
    \includegraphics[width=13cm]{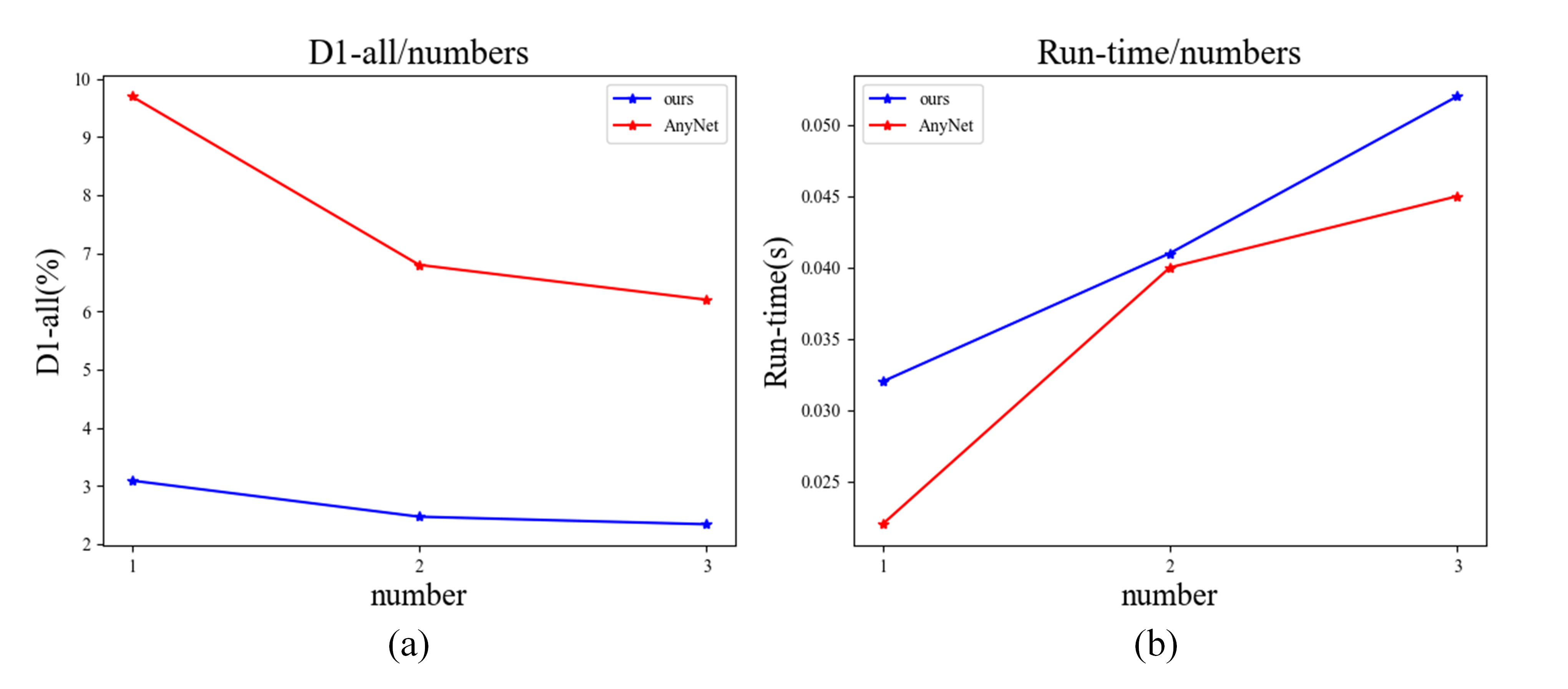}
    \caption{Impact of different numbers of stacks on network performance metrics. (a) Change in the accuracy evaluation metric D1-all. (b) Change in the computational time evaluation metric Run-time.}
    \label{fig:13}
\end{figure}

In the design of the AnyNet structure, a similar stacking concept is employed. However, RecSM exhibits superior accuracy performance, as demonstrated in the comparative results shown in Figure \ref{fig:13}.

\subsubsection{Analysis of DOM}
As a functionally independent module, DOM plays a role in optimizing disparity output. The experimental comparison on whether to use DOM in SCS is shown in the Table \ref{tab:5}.
\begin{table}[htbp]
\centering
\caption{The impact of using DOM on network performance when SCS stacks 1 and 3 separately.}
\begin{tabular}{cc|ccc}
\hline
Number of SCS & The structure of SCS & EpE(px) & D1-all(\%) & Run-time(s) \\
\hline
 1 & only MREM & 1.036 & 5.305 & 0.030 \\
 1 & MREM+DOM & 0.802 & 3.082 & 0.032 \\
\hline
 3 & only MREM & 0.938 & 4.452 & 0.046 \\
 3 & MREM+DOM & 0.737 & 2.336 & 0.051 \\
\hline
\end{tabular}
\label{tab:5}
\end{table}

The experimental results regarding the "shared-weight" factor are presented in Table \ref{tab:6}.
\begin{table}[htbp]
\centering
\caption{Comparative results of experimental metrics regarding the shared-weight factor in the DOM}
\begin{tabular}{c|ccc}
\hline
Shared-Weight & EpE(px) & D1-all(\%) & Run-time(s) \\
\hline
  & 1.01 & 4.97 & 0.08 \\
\checkmark & \textbf{0.737} & \textbf{2.336} & \textbf{0.054} \\
\hline
\end{tabular}
\label{tab:6}
\end{table}

The use of the shared-weight DOM results in improvements across various aspects of the network's performance. It appears to be more robust compared to employing independently separate DOM.

\subsection{RecSM Performance}
The performance comparison results of RecSM and existing mainstream algorithms on the KITTI dataset are shown in Table \ref{tab:7}.

\begin{table}[htbp]
\centering
\caption{Performance comparison of KITTI stereo evaluation in the same experimental environment. (* means this indicator is not available)}
\begin{tabular}{c|cc|cc|c}
\hline
 & \multicolumn{2}{|c|}{KITTI 2012} & \multicolumn{2}{c|}{KITTI 2015} \\
\hline
Method & EpE(px) & 3px-all(\%) & EpE(px) & D1-all(\%) & Run-time(s) \\
\hline
 MC-CNN \citep{zbontar2016stereo} & 0.9 & 3.50 & * & 3.89 & 67 \\
 ELFNet \citep{lou2023elfnet} & 1.18 & 4.74 & 1.57 & 5.82 & * \\
 GANet-deep \citep{zhang2019ga} & \textbf{0.5} & 1.60 & * & 1.81 & 1.8 \\
 GC-Net \citep{kendall2017end} & 0.7 & 2.30 & 2.51 & 2.87 & 0.9 \\
 SSPCV-Net \citep{wu2019semantic} & 0.6 & 1.9 & 1.84 & 2.11 & 0.9 \\
 Segstereo \citep{yang2018segstereo} & 0.6 & 2.03 & 1.84 & 4.83 & 0.6 \\
 AcfNet \citep{zhang2020adaptive} & \textbf{0.5} & 1.54 & * & 1.89 & 0.48 \\
 PSMNet \citep{chang2018pyramid} & 0.6 & 1.89 & 1.09 & 2.32 & 0.41 \\
 GwcNet \citep{guo2019group} & \textbf{0.5} & 1.70 & * & 2.11 & 0.32 \\
 ACVNet \citep{xu2022attention} & \textbf{0.5} & \textbf{1.47} & \textbf{0.5} & \textbf{1.65} & 0.2 \\
 CFNet \citep{shen2021cfnet} & \textbf{0.5} & 1.58 & * & 1.88 & 0.18 \\
 TANet \citep{zhao2023fast}& \textbf{0.5} & 1.57  & 0.81 & 2.71 & 0.061 \\
 RecSM(3 SCSs) & 0.6 & 1.61 & 0.74 & 2.34 & \textbf{0.054} \\
\hline
\end{tabular}
\label{tab:7}
\end{table}

In comparison to existing algorithms, RecSM demonstrates a significant advantage in terms of computational speed. It achieves faster processing while maintaining accuracy levels comparable to algorithms within the same performance range. Furthermore, RecSM possesses the flexibility of network structure that other algorithms do not, effectively striking a balance between speed and accuracy in algorithmic performance.
\label{sec:Experiment}

\section{Conclusion}
In response to the challenge of slow stereo matching algorithms, We propose a flexible recursive network for stereo matching based on residual estimation which abbreviated as RecSM. RecSM leverages temporal context and incorporates temporal attention while employing residual estimation to reduce the search range, thus mitigating computational demands. Additionally, we have designed a flexible stackable computation structure tailored to the characteristics of residual estimation. By SCS, we continuously optimize disparity results, enhancing the network's versatility. 

Nevertheless, RecSM still offers substantial room for improvement. For instance, disparities of objects at close range typically exhibit larger variations, making them challenging to match with fixed search range algorithms. Further error reduction may necessitate post-processing techniques like disparity refinement. Similarly, based on RecSM's algorithmic principles, higher camera frame rates result in fewer disparity change regions between adjacent frames, leading to reduced matching difficulties. However, RecSM's performance may be constrained when dealing with low frame rate scenarios.

In the future, the way RecSM incorporates temporal context can be further optimized. It could involve adaptive partitioning of temporal context, distinguishing between occluded regions and non-co-visible areas, and making separate adjustments to reduce the number of mismatched pixels. Additionally, during the aggregation process of the residual cost volume, we currently employ a straightforward stacking of 3D convolutions. However, there may be more suitable computational methods that warrant experimental exploration. Similarly, there is still room for exploration in the selection of backbone. We attempted to use tiny types of Swin-Transformer \citep{liu2021swin} as backbone, but the results were mediocre, with D1-all only reaching 2.62\%. However, this indicates the possibility of better backbone.  Considering real-world road scenarios, integrating dynamically calibrated results into RecSM, especially for pitch angle correction, may also potentially reduce matching difficulties.
\label{sec:Conclusion}

 \bibliographystyle{elsarticle-num} 

 \bibliography{ref.bib}




\end{document}